\documentclass[runningheads]{llncs}
\usepackage{graphicx}
\usepackage{comment}
\usepackage{amsmath,amssymb} %
\usepackage{color}
\usepackage{amssymb}
\usepackage{amsmath}
\usepackage{stmaryrd}
\usepackage{textcomp}
\usepackage{booktabs} 

\newcommand{\bp}{\textbf{p}} %

\newcommand{\cL}{\mathcal{L}} %

\def\etal{\emph{et al}}
\def\ie{\emph{i.e.}}

\newcommand{\mv}[1]{\mathbf{#1}}

\newcommand{\ilf}{\mv{L}_{1}}
\newcommand{\ils}{\mv{L}_{2}}
\newcommand{\irf}{\mv{R}_{1}}
\newcommand{\irs}{\mv{R}_{2}}
\newcommand{\il}{\mv{L}}
\newcommand{\ir}{\mv{R}}

\newcommand{\of}{\mv{O}_f}

\newcommand{\fl}{\mv{F}_{1 \rightarrow 2}}
\newcommand{\flb}{\hat{\mv{F}}_{2 \rightarrow 1}}

\newcommand{\df}{\mv{D}_1}
\newcommand{\ds}{\mv{D}_2}
\newcommand{\dc}{\mv{C}_{1 \rightarrow 2}}

\newcommand{\sfl}{\mv{X}}
\newcommand{\grad}{\nabla\cL}

\newcommand{\ours}{\textit{CGSF}}

\newcommand{\init}{\mv{H}_\theta}
\newcommand{\refi}{\mv{G}_\alpha}

\DeclareMathOperator*{\argmin}{arg\,min}

\usepackage[width=122mm,left=12mm,paperwidth=146mm,height=193mm,top=12mm,paperheight=217mm]{geometry}

\begin{document}

\pagestyle{headings}
\mainmatter
\def\ECCVSubNumber{}  
\title{Consistency Guided Scene Flow Estimation}

\title{Consistency Guided Scene Flow Estimation} 

\titlerunning{Consistency Guided Scene Flow Estimation}

\author{Yuhua Chen\inst{1,2} \and
Luc Van Gool\inst{2} \and
Cordelia Schmid\inst{1} \and 
Cristian Sminchisescu\inst{1}}
\authorrunning{Y. Chen et al.}

\institute{Google Research \and
ETH Zurich \\}

\newcommand{\todo}[1]{\textbf{\textcolor{red}{[#1]}}}

\maketitle

\begin{abstract}

Consistency Guided Scene Flow Estimation (\ours) is a self-supervised framework for the joint reconstruction of 3D scene structure and motion from stereo video. The model takes two temporal stereo pairs as input, and predicts disparity and scene flow. The model self-adapts at test time by iteratively refining its predictions. The refinement process is guided by a consistency loss, which combines stereo and temporal photo-consistency with a geometric term that couples disparity and 3D motion. To handle inherent modeling error in the consistency loss (e.g. Lambertian assumptions) and for better generalization, we further introduce a learned, output refinement network, which takes the initial predictions, the loss, and the gradient as input, and efficiently predicts a correlated output update. In multiple experiments, including ablation studies, we show that the proposed model can reliably predict disparity and scene flow in challenging imagery, achieves better generalization than the state-of-the-art, and adapts quickly and robustly to unseen domains. 

\end{abstract}

\section{Introduction}
Scene flow is the task of jointly estimating the 3D structure and motion of a scene~\cite{vedula1999three}. This is critical for a wide range of downstream applications, such as robotics, autonomous driving and augmented reality. Typically, scene flow is estimated from consecutive stereo pairs, and requires simultaneously solving two sub-problems: stereo matching and optical flow. Though closely related, the two tasks cover different aspects of scene flow: stereo matching seeks to estimate disparity from a static stereo pair, whereas the objective of optical flow is to find the correspondence between temporally adjacent frames. 

Many end-to-end models powered by deep neural network have emerged for the scene flow estimation~\cite{aleotti2020learning,jiang2019sense} and its components~\cite{chang18pyramid,ilg2017flownet,sun2018pwc}, and they showed promise in benchmarks. However, training such models relies on the availability of labeled data. Due to difficulty in acquiring real-world data, synthetic data~\cite{mayer16a} is widely used as the major supervision source for deep models, with 3D structure or motion generated automatically using computer graphics techniques. Unfortunately, synthetic data still exhibits noticeable appearance dissimilarity leading to domain shift, which often prevents models trained this way from generalizing to the real-world. This has been observed previously~\cite{pang2018zoom,tonioni2017unsupervised}. This issue can be partially addressed by finetuning on labeled real-world data. However, collecting ground-truth labels for scene flow can be extremely challenging, requiring the use of costly sensors and additional manual intervention~\cite{Menze2015ISA,Menze2018JPRS}. This, however, may not always be feasible in many real-world scenarios, greatly limiting the applicability of deep models.

To address this issue, we present Consistency Guided Scene Flow (\ours), a framework that additionally models output consistency. The framework begins with producing scene flow prediction using feedforward deep network. The predictions include disparity, optical flow and disparity change. These predictions are then coupled by a consistency loss which captures the photometric and geometric relations valid in scene flow, irrespective of domain shift.
Therefore, the consistency loss can be a powerful cue in guiding predictions for better generalization in unseen domains. However, the consistency loss is inherently noisy due to the complexity of natural data, such as non-Lambertian surfaces or occlusion. As a result, the provided guidance is not always precise, and can exhibit undesired artifacts. To further correct such issues that are difficult to model explicitly, we further introduce a learned refinement module, which takes the initial predictions, the loss, and the gradient as input, and predicts an update to recursively refine the prediction. The refinement module is implemented as a neural network, and thus can be trained jointly with the original feedforward module. 

Our \ours{} model can be trained either using synthetic data, which can generalize well to real imagery, or trained self-supervised, using unlabeled data, based on the proposed consistency loss. In diverse experiments, we show our \ours{} can reliably predict disparity and scene flow in challenging scenarios. Moreover, we observe that the proposed, learned refinement module, can significantly improve the results of the more classical feedforward network, by ensuring consistent predictions. In particular, we demonstrate that the proposed model significantly improves generalization to unseen domains, thus better supporting real-world applications of scene flow, where a degree of domain shift is inevitable. 

\section{Related Work}
\noindent{\bf Scene Flow Estimation.} introduced by Vedula \etal.~\cite{vedula1999three}, scene flow estimation aims to recover the 3D structure and motion of a scene. The task has been formulated as a variational inference problem by multiple authors~\cite{basha2013multi,huguet2007variational,vogel2013piecewise,wedel2008efficient}. Recently, several deep models have emerged for this task. Ilg \etal.~\cite{Ilg2018occlusions} combine networks for stereo and optical flow using an additional network to predict disparity change. Ma \etal.~\cite{Ma2019DRISF} stack three networks to predicting disparity, flow and segmentation, then use a Gaussian-Newton solver to estimate per-object 3D motion. To leverage correlation between tasks, Jiang \etal.~\cite{jiang2019sense} propose an encoder architecture shared among the tasks of disparity, optical flow, and segmentation. Similarly, Aleotti \etal.~\cite{aleotti2020learning} propose a lightweight architecture to share information between tasks. Complementary to previous work, our main objective is to improve generalization of the scene flow by means of additional constraints, self-supervised losses, and learnt refinement schemes. 

\noindent{\bf Flow and Disparity Estimation.} 
As scene flow requires simultaneously reasoning about two tasks: disparity and flow estimation, it has been largely influenced by the techniques used in end-to-end stereo matching and optical flow. 

For optical flow estimation, FlowNet~\cite{dosovitskiy2015flownet} represents the first attempt in building end-to-end deep models. FlowNet 2.0~\cite{ilg2017flownet} further improved performance by dataset scheduling and refinement.  
A spatial pyramid design is adopted in PWC-Net~\cite{sun2018pwc}, which includes cost volume processing at multiple levels. PWC-Net achieved highly competitive performance on several benchmarks for optical flow estimation.

In stereo matching, Mayer \etal.~\cite{mayer16a} introduce an architecture similar to FlowNet, for disparity estimation from rectified image pairs. Several techniques have been introduced to improve the matching accuracy, including 3D convolution in GC-Net~\cite{Kendall_2017_ICCV}, pyramid pooling in PSM-Net~\cite{chang18pyramid}, \textit{etc}.  More recently, GA-Net~\cite{zhang2019ga} integrated two aggregation layers with good results.

\noindent{\bf Adaptation.} Due to the practical difficulty of acquiring ground-truth labels, the aforementioned models are often trained on synthetic data, although a performance drop is observed when applied in the real-world~\cite{pang2018zoom,tonioni2017unsupervised}.

To address this issue, several techniques~\cite{guo2018learning,poggi2019guided,tonioni2017unsupervised,tonioni2019learning,tonioni2019real,zhong2018open} have been proposed to adapt the model to new domains where labels aren't available. For the purpose, either offline adaptation~\cite{tonioni2017unsupervised} or online adaptation~\cite{tonioni2019real} is performed to refine the model parameters in the new domain. One key difference in our work is that our refinement module operates on prediction (outputs) directly instead of network parameters, and it thus relies on the output consistency among outputs to overcome domain shift.

\noindent{\bf Self-supervision.} On the other hand, self-supervised learning has been used with unlabelled data~\cite{casser2019depth,godard2017unsupervised,godard2019digging,lai2019bridging,wang2019unos,yin2018geonet,zhou2017unsupervised}. Some methods require collecting unlabelled data for offline training, which might not be always possible. Even so, trained model suffer from domain shift in new environments, which still requires online adaptation. Part of our modeling is based on similar ideas, in that we can also train with self-supervision in an offline stage. However, we additionally integrate output consistency terms in our refinement module, which makes our framework easier to self-adapt to new environments without collecting unlabeled data upfront.

\noindent{\bf Online Refinement.} Online optimization under self-supervised consistency losses has shown to be effective for depth prediction in video~\cite{casser2019depth,chen2019self}. In this work, we learn a refinement network to preserve the output consistency. Our refinement network shares ideas with learning-based optimization~\cite{andrychowicz2016learning,li2016learning}, where one aims to integrate iterative refinement into deep networks. Different metalearning ideas have also been explored in different applications, such as learning depth and pose estimation~\cite{clark2018ls}, 3D rigid motion estimation~\cite{lv2019taking}, or monocular reconstruction of static scenes~\cite{tang2018ba}. Here we focus on problem-domain updates that integrate gradient information in order to ensure progress, for robustness, as well as computational efficiency.
In contrast, our model consists of several novel geometric constraints for scene flow estimation, designed for self-supervision and overcoming domain shift.

\section{Consistency Guided Scene Flow}
\subsection{Overview}
\begin{figure*}
\vspace{-5mm}
\includegraphics[trim=0 400 400 0, clip, width=\linewidth]{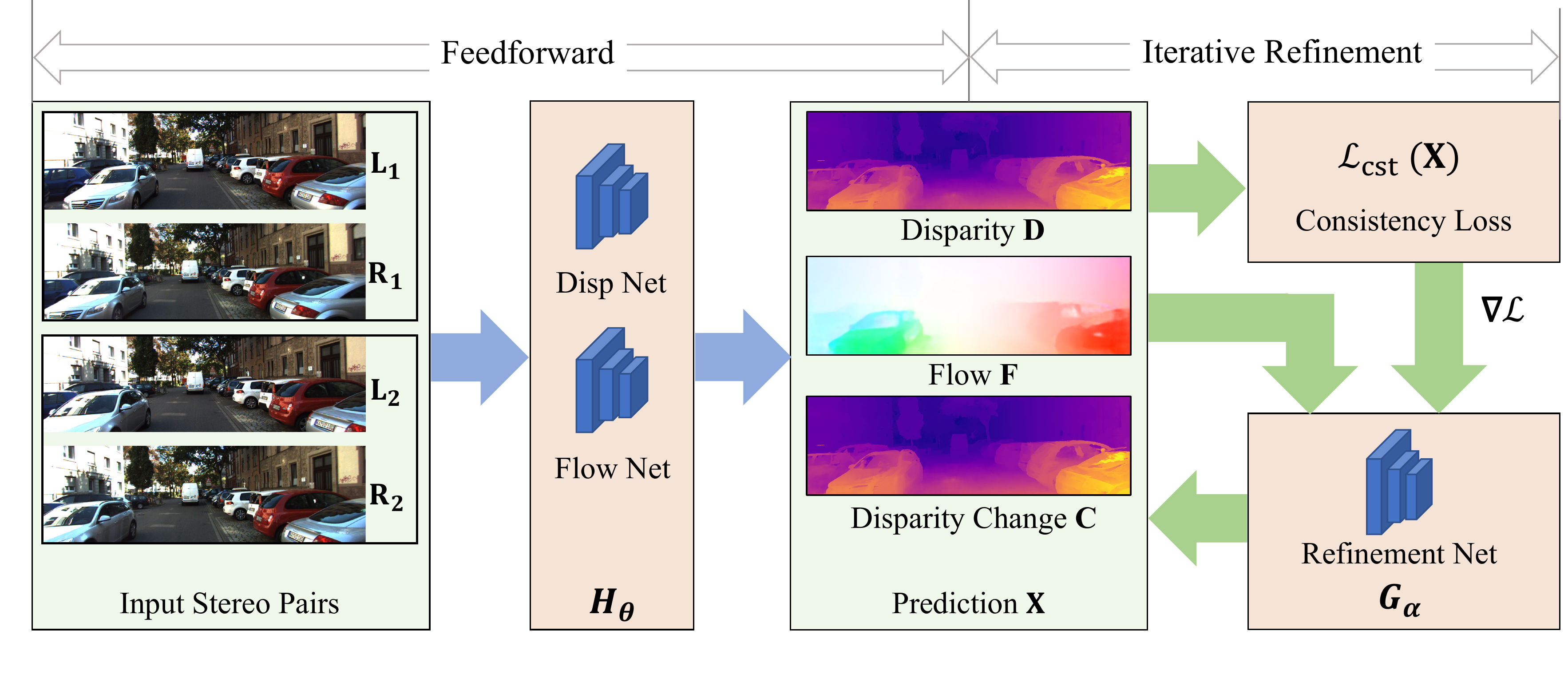}

\caption{\textbf{Illustration of our proposed Consistency Guided Scene Flow (\ours)}. The model can be obtained either using labeled data, or unlabeled images and self-supervision. At runtime, it self-adapts by producing a set of outputs using feedforward networks ($\init$, flow by blue arrows), then iteratively refining those estimates on two temporally adjacent stereo pairs, using a learned update (network $\refi$, flow by green arrows). The refinement process is guided by a consistency loss, which captures several intrinsic photometric and geometric constraints associated to scene flow.}
\label{fig:overview}
\end{figure*}

In this section we present our Consistency Guided Scene Flow framework (\ours). An overview is given in fig.~\ref{fig:overview}.
Given a couple of stereo images at subsequent timesteps $\left(\ilf, \irf\right)$ and $\left(\ils, \irs\right)$, respectively, as input, the model predicts scene flow $\sfl$, \ie, the 3D structure and motion of the environment. The 3D scene structure can be represented by the disparity in the first frame $\df$, as disparity is inversely proportional to depth. The 3D scene motion can be decomposed into the motion parallel to the image plane, represented by optical flow $\fl$, and the motion orthogonal to the image plane, represented as disparity change $\dc$.

At runtime, we formulate scene flow estimation as an iterative optimization problem. Our framework predicts scene flow in two stages: a more typical feedforward block and an iterative output refinement module (N.B. this is also feedforward, but has a recurrent structure). In the first stage, we rely on a feedforward module $\init$ to predict scene flow given input images. In the refinement stage, our geometric and photometric consistency losses are further optimized with respect to the outputs using a refinement network $\refi$. Both the feedforward module $\init$ and the refinement network $\refi$ are implemented as neural networks with learnt parameters $\theta$ and $\alpha$. Therefore the two components can be trained jointly in an end-to-end process.

In the sequel, we introduce the feedforward module in \S\ref{sec:init}. We formulate the geometric and photometric consistency terms for scene flow estimation in \S\ref{sec:loss}. We present the learned refinement network in \S\ref{sec:refine} and describe the training and evaluation protocol in \S\ref{sec:training}.

\subsection{Feedforward Scene Flow Module}
\label{sec:init}

In the feedforward stage, the prediction of scene flow is obtained by passing the input images through disparity and flow networks. The disparity network predicts the 3D structure given stereo pairs. Specifically, the module predicts $\df$ from $\left(\ilf, \irf\right)$, and $\ds$ from $\left(\ils, \irs\right)$. %
Given $\left(\ilf, \ils\right)$ as input, the flow network predicts optical flow $\fl$, and the disparity change $\dc$ which encodes the motion component orthogonal to the image plane. 

In this work, we build the feedforward module upon GA-Net~\cite{zhang2019ga} and PWC-Net~\cite{sun2018pwc}, mainly due to both competitive performance and implementation compatibility. Here we briefly introduce the architectures.

\noindent\textbf{Disparity network}: We rely on GA-Net~\cite{zhang2019ga} to predict scene structure, \ie, $\df$ and $\ds$. The network extracts features from a stereo pair with a shared encoder. Then features are used to build a cost volume, which is fed into a cost aggregation block for regularization, refinement and disparity prediction.

\noindent\textbf{Flow network}: We modify PWC-Net~\cite{sun2018pwc} to predict the 3D motion, including optical flow $\fl$ and disparity change $\dc$. We follow the design in PWC-Net to construct a cost volume from the features of the first frame, and the warped features of the second frame. The cost volume is further processed to obtain the motion prediction. We increase the output channel from $2$ to $3$ to encode 3D motion.

In principle, our framework is agnostic to specific choices of model design--other options for the structure and motion networks are possible. This offers the flexibility of leveraging the latest advances in the field, and focus on our novel self-supervised geometric and photometric consistency losses, described next. 

\subsection{Scene Flow Consistency Loss}
\label{sec:loss}

In this section we formulate the photometric and geometric constrains that are part of our proposed consistency loss.

\vspace{-2mm}
\subsubsection{Stereo Photometric Consistency.}

Photometric consistency between stereo pairs is widely used as a self-supervised loss in deep models~\cite{godard2017unsupervised,kuznietsov2017semi}. The idea is to warp the source view via a differentiable bilinear map produced by the network prediction. The photometric difference between the warped source image and the target image can be used as a self-supervision signal. 

Given a stereo pair $(\il, \ir)$, we use the left view $\il$ as the target image, and the right view $\ir$ as the source image. Considering a pixel $\bp$ in the left view, its correspondence in the right view can be obtained by subtracting its disparity value from its x coordinate. With a slight abuse of notation, we denote it as $\bp - \mv{D}(\bp)$. If $\bp$ corresponds to a 3D scene point visible in both views, its correspondence should be visually consistent. Following \cite{godard2017unsupervised}, we use a weighted sum of L1 loss and SSIM~\cite{wang2004image} loss as the photometric error, denoted as $pe$. Therefore the photometric stereo consistency model can be written as

\begin{align}
    \cL_{pd}(\bp) =  pe(\il(\bp), \ir(\bp - \mv{D}(\bp)))
\end{align}
The stereo consistency applies to both $\df$ and $\ds$.
\subsubsection{Flow Photometric Consistency.} 
We also rely on photometric consistency for temporal frames. Given a pixel $\bp$ in the first frame $\ilf$, its correspondence in the second frame $\ils$ can be obtained by displacing $\bp$ using optical flow. We can write the flow photometric consistency as
\begin{align}
    \cL_{pf}(\bp) = \of(\bp) \cdot pe(\il(\bp), \il(\bp + \fl(\bp)))
\end{align}
\noindent $\of$ represents the occlusion map caused by optical flow. Similarly with previous work~\cite{liu2019selflow,meister2018unflow}, we infer the occlusion map using a forward-backward flow consistency check. In more detail, an additional backward flow is computed and then warped by the the forward flow $\fl$. We denote the computed warped backward flow map by $\flb$. A forward-backward consistency check is performed to estimate the occlusion map
\begin{align}
    \of = \llbracket |\fl + \flb|^2 < w_1 (|\fl|^2 + |\flb|^2) + w_2  \rrbracket
\end{align}
\noindent where $\llbracket \cdot \rrbracket$ is the Iverson bracket, and we set $w_1=0.01, w_2=0.05$ in inferring the occlusion mask.

\subsubsection{Disparity-Flow Consistency.} 
We formulate a new consistency term to connect disparity and 3D motion. Given a pixel $\bp$ in the left view of the first frame $\ilf$, its correspondence in $\ils$ is associated by optical flow $\fl$ can be written as $\bp + \fl(\bp)$. Thus its disparity value in the second frame should be $\ds(\bp + \fl(\bp))$. On the other hand, $\bp$'s disparity value in the second frame can also be obtained by adding the disparity change $\dc$ to the disparity of the first frame $\df$. Therefore the disparity flow consistency can be written as
\begin{align}
    \cL_{df}(\bp) = \of(\bp) ||\df(\bp) + \dc(\bp) - \ds(\bp + \fl(\bp))|| 
\end{align}
\noindent We use $\of$ to mask out occluded pixels as flow warping is used in the process.

\subsubsection{Overall Consistency Loss. }
By integrating all previously derived terms, the consistency loss can be written as
\begin{align}
    \cL_{cst} = \cL_{pd} + \cL_{pf} + \cL_{df}
\end{align}
The consistency loss regularizes predictions resulting from the individual structure and flow networks $\sfl$, and captures several intrinsic geometric and photometric constraints that should always hold. To provide additional regularization of predictions, we use an edge-aware smoothness term as in \cite{godard2017unsupervised}. This is applied to the disparity map $\df, \ds$, optical flow $\fl$, and disparity change $\dc$.

\subsection{Consistency Guided Refinement}
\label{sec:refine}

In the refinement stage, the goal is to recursively improve scene flow prediction. We denote as $\sfl^t$ the scene flow prediction after $t$ steps of refinement. The initial prediction from the classical feedforward module can be denoted as $\sfl^0$. 

As the consistency loss $\cL_{cst}$ reflects the intrinsic structure of scene flow, it can be used as an indicator of how good the prediction is. Therefore we aim to leverage the signal in consistency loss to guide online finetuning. To this end, we build a refinement network $\refi$ parameterized by $\alpha$, which takes as input the previous prediction $\sfl^{t}$, the consistency loss over the previous prediction $\cL_{cst}(\sfl^{t})$, and the gradient on predictions $\nabla _{\sfl}\cL_{cst}(\sfl^{t})$. The reason for using the gradient is to provide an exact signal for the descent direction. The refinement network then predicts an update to refine the prediction to $\sfl^{t+1}$. 
\begin{align}
    {\sfl}^{t+1} = {\sfl}^{t} + \refi(\sfl, \cL_{cst} ,\nabla_{\sfl} \cL_{cst})
\end{align}
For implementation, we first concatenate all predictions, including $\df$, $\ds$, $\dc$ each with size $W\times H\times 1$, and optical flow $\fl$ with size $W\times H\times 2$. The concatenated prediction map is of dimension $W \times H\times 5$, with $W$ and $H$ being the width and height of the image, respectively. The gradient is of the same dimensionality as the prediction, and the loss map is of $1$ channel. Therefore, all inputs can be concatenated as a $W\times H\times 11$ map. The benefit of using a concatenated map as input is that cross-channel correlations can be better captured. For instance, disparity information might be helpful to improve optical flow. 

We implement the refinement module as a small fully convolutional network(FCN) operating on the input map, to provide dense (per pixel) output for the updates. The advantage is that the FCN architecture can learn to leverage the local inter-pixel context, which is helpful in reducing noise in the consistency loss, and in propagating information to and from neighbouring pixels. We use a lightweight design of $3$ convolution layers, each with $512$ hidden units and kernel size of $3\times3$. ReLU is used between convolution layers as the activation function. The light-weight design is motivated by the recurrent use of the module. 
\subsection{Training}

\label{sec:training}
To learn the parameters of the feedforward scene flow $\init$ and refinement network $\refi$, the framework can be initialized based on pre-training using synthetic data with ground-truth labels (whenever available), and then/or further trained using the self-supervised loss $\cL_{cst}$, in order to better generalize to real imagery. When ground-truth labels are available, a standard supervised loss can be used
\begin{align}
    \cL_{sup}(\sfl) = ||\fl - \fl^*||  + ||\df - \df^*|| + ||\dc - \dc^*|| 
\end{align}
\noindent where $*$ indicates ground-truth.
In either case, the network parameters $\theta,\alpha$ can be learned jointly by minimizing the loss using stochastic gradient descent. The optimization objective writes
\begin{align}
    {\argmin_{\theta, \alpha}} \sum_{t\in\{0,...,T\}} \cL(\sfl^t)
\end{align}
\noindent where the loss $\cL$ can be either $\cL_{sup}$(when the label is available) or $\cL_{cst}$(in self-supervised training). 
The loss is applied to the initial prediction, as well as to the prediction after each refinement step (i.e. we supervise the augmented step update at each iteration). Essentially after each step we minimize the loss w.r.t. $\boldsymbol{\alpha}$ so that the learnt descent direction produces as large of a loss decrease as possible, and aggregate in parallel over $T=5$ refinement steps.

\section{Experiments}

In this section, we experimentally validate the proposed \ours. First, we test our model on synthetic data on FlyingThings3D in \S\ref{sec:fly}. Then, in order to verify real-image generalization, we test the model on KITTI in \S\ref{sec:kitti}. In \S\ref{sec:ablate} we provide ablation studies to provide further insight into our design choices. Finally, in \S\ref{sec:zed} we show the performance of \ours{} on different scenes captured by a stereo camera, in order to demonstrate the generalization to environments not seen in training.

\subsection{Evaluation on Synthetic Data}
\label{sec:fly}

First we evaluate the model on synthetic data. We use the FlyingThings3D~\cite{mayer16a} which is a synthetic dataset for scene flow estimation, containing $22,390$ stereo training pairs. Dense ground-truth labels are available in all images for disparity, optical flow and disparity change. We use the ground-truth to supervise our model, including the feedforward module and the refinement network. 

In the refinement module, we use a set of trade-off weights to balance different consistency terms. We use $\omega_{pd}, \omega_{pf}, \omega_{df}$ to denote, respectively, the weight of stereo photometric consistency $\cL_{pd}$, flow disparity consistency $\cL_{pf}$, and disparity disparity-flow consistency $\cL_{df}$. The weight of the smoothness term is denoted as $\omega_{s}$. We set $\omega_{pd} = 1, \omega_{pf}=1, \omega_{df} = 1, \omega_{s} = 0.1$. A stage-wise strategy is used in order to improve the stability of the training process. In more detail, we first train the disparity network and the flow network separately. For this part, we follow \cite{zhang2019ga} and \cite{sun2018pwc} in setting hyperparameters. These are then used as initialization for additional, joint fine-tuning. Specifically, we randomly initialize the weights of the refinement module, and jointly fine-tune the whole network including the pre-trained feedforward networks. To fine-tune the entire network, we use the Adam optimizer with $\beta_1 = 0.9$ and $\beta_2 = 0.999$. The learning rate is set to $10^{-4}$. The network is trained on FlyingThings 3D for 10 epochs with a batch size of $4$.

The models are evaluated on the test set of FlyingThings3D, which contains $4,370$ images. As scene flow model requires $4$ images as input, we drop the last frame of every sequence, resulting in a test set with $3,933$ images. The performance is measured by end point error(EPE) for the three key outputs: optical flow, disparity, and disparity change. 
We use the results produced by the feedforward module without a refinement network as baseline. To facilitate the comparison with the state-of-the-art, we test against a recent model DWARF~\cite{aleotti2020learning} in the same experimental setting. As shown in table~\ref{tab:sf_fly}, the refinement model improves over the feedforward baseline in all three metrics, thus supporting the effectiveness of the consistency guided refinement process. Our model also outperforms DWARF~\cite{aleotti2020learning} by a large margin. Again, this shows the benefit of geometrically modelling the consistency of scene flow outputs. 

\begin{table}[t]
\begin{center}
{
\setlength{\tabcolsep}{5.0pt}
\begin{tabular}{c  c  c  c}
\toprule
  & Disparity EPE  & Flow EPE & Disparity Change EPE * \\ \midrule
DWARF~\cite{aleotti2020learning} & 2.00  & 6.52 & 2.23 \\ 
Feedforward & 0.83 & 7.02 & 2.02  \\
\ours (ours) & \textbf{0.79}  & \textbf{5.98} & \textbf{1.33} \\ \bottomrule

\end{tabular}
}
\end{center}
\caption{\textbf{Scene flow results on the FlyingThings3D test set}. Results are evaluated as end point error (EPE). All models are trained on the training set of FlyingThings3D. Disparity change EPE represents the error of the motion orthogonal to the image plane.}
\label{tab:sf_fly}
\end{table}

\subsection{Generalization to Real Images}

\label{sec:kitti}
\begin{table}[t]
\begin{center}
\begin{center}
{
\setlength{\tabcolsep}{5.0pt}
\begin{tabular}{c c c  c  c  c}
\toprule
Method & Training Data & D1 & D2 & F1 & SF \\ \midrule
Godard \etal~\cite{godard2017unsupervised} & K(u)  
 &   9.19 & - & -  & -\\  
BridgeDepthFlow~\cite{lai2019bridging} &K(u)  
 &   8.62 & - & 26.72 & -\\ 
UnOS~\cite{wang2019unos} &K(u)  
 &   5.94 & - & 16.30 & -\\ 
MADNet~\cite{tonioni2019real}  &F(s)+ K(u)  
 &   8.41 & - & - & -\\ 
DWARF~\cite{aleotti2020learning} &F(s)
& 11.60  & 29.64 & 38.32 & 45.58 \\  \hline

Feedforward &F(s)
& 11.53 & 27.83  & 33.09 & 43.55 \\  
\ours (ours) & F(s)
& 7.10  & 19.68  & 27.35 &  35.40  \\ 
\ours (ours) & K(u) 
& 6.65 & 17.69 & 23.05 & 31.52  \\ 
\ours (ours) &F(s)+ K(u) 
& 5.75 & 15.53 & 20.45 & 28.14  \\ \bottomrule

\end{tabular}
}
\end{center}
\end{center}
\caption{\textbf{Evaluation on the KITTI 2015 scene flow dataset.} The percentage of outliers is used as evaluation metric. A pixel is considered as correct if the prediction end-point error is smaller than $3px$ or $5\%$. For training data, 'K(u)' stands for unsupervised training on KITTI, and 'F(s)' represents supervised training on FlyingThings3D. }

\label{tab:sf_kitti}
\end{table}

To verify the performance in real imagery where no labeled data is available, we evaluate on the KITTI scene flow dataset~\cite{Menze2015ISA,Menze2018JPRS}, which contains 200 training images with sparse ground-truth disparity and flow acquired by a lidar sensor. The ground-truth is only used in evaluation. To comply with the KITTI scene flow benchmark, the percentage of outliers in the following $4$ categories is used as evaluation metrics, D1: outliers in the disparity of first frame (\ie, $\df$), D2: outliers in second frame warped to the first frame (\ie, $\df + \dc$), F1: outliers in optical flow (\ie, $\fl$), SF: outliers in either D1,D2 or F1. A pixel is considered correct if the prediction end-point error is smaller than $3px$ or $5\%$.

As discussed in \S\ref{sec:loss}, our \ours{} model can be trained with round-truth labels from synthetic data, or using self-supervision. We evaluate both strategies. For training using ground-truth on synthetic data, we already illustrated FlyingThings3D. For self-supervised training on KITTI, we use KITTI raw as a training set, with all test scenes excluded. 

We compare with the supervised feedforward network without refinement as a baseline. Additionally, we compare with several recent competing methods, including self-supervised models based on stereo images~\cite{godard2017unsupervised} or stereo video~\cite{lai2019bridging,wang2019unos}, scene flow supervised on synthetic data~\cite{aleotti2020learning}, and an adaptation model for stereo estimation~\cite{tonioni2019real}. We initialize \cite{tonioni2019real} with the pre-trained weights from synthetic data, and do unsupervised adaptation on KITTI raw (the same set used in our model's unsupervised training).

The results are summarized in table~\ref{tab:sf_kitti}. With the feedforward module only, the baseline achieves an error rate of $43.55\%$. The error can be significantly reduced to $35.40\%$ by the proposed refinement network. Notably, the performance gain is much larger compared to the one in FlyingThings3D, which demonstrates that consistency plays a central and positive role in cross-domain scenarios. This indicates that consistency can be used as an effective model to overcome domain gaps and improve generalization. 

Whenever collecting an unlabeled training set beforehand is possible, the model can also be trained under self-supervision. This results in an error rate of $31.52\%$, which demonstrates both the effectiveness of a self-supervised consistency loss, and good compatibility when supervision is available. Performance can be further improved by combining pre-training on synthetic data, and self-supervised finetuning. This combination results in the lowest scene flow error rate of $28.14\%$. 

Compared to other component methods, our model outperforms the scene flow approach DWARF~\cite{aleotti2020learning}. The performance of disparity estimation is also considerably better than the self-adaptive stereo model MADNet~\cite{tonioni2019real}. Compared to other self-supervised models relying on stereo data~\cite{godard2017unsupervised,lai2019bridging,wang2019unos}, our model achieves competitive results on disparity and flow estimation, while being additionally capable of predicting 3D motion.

\begin{table}[t]
\begin{center}
\setlength{\tabcolsep}{5.0pt}
\begin{tabular}{c c  c  c c  c  c  c}
\toprule
Refinement & $\cL_{pd}$ & $\cL_{pf}$ & $\cL_{df}$ &  D1 & D2 & F1 & SF  \\ \midrule
& & & & 
11.29 & 26.51 & 32.71 & 43.01 \\ 
\checkmark & & & & 
10.40 & 25.15 & 32.05 & 42.43 \\ 
\checkmark & \checkmark &  &  & 
7.95  & 22.48 & 31.38 & 39.80 \\ 
\checkmark & & \checkmark &  & 
9.84  & 23.92 & 29.06 &  39.64 \\ 
\checkmark & &  & \checkmark & 
 9.11  & 21.14 & 30.95 & 39.59 \\ 
\checkmark & \checkmark & \checkmark &  & 
7.60  & 22.73 & 29.07 & 38.23 \\ 
\checkmark & \checkmark & \checkmark & \checkmark & 
7.10  & 19.68  & 27.35 &  35.40 \\ \bottomrule
\end{tabular}
\end{center}
\caption{\textbf{Ablation study for the consistency terms.} Different consistency terms are used in the refinement module. All models are trained on FlyingThings3D and evaluated on KITTI.}
\label{tab:ablation_loss}

\end{table}

\begin{figure*}[t]
\small
\centering
\input{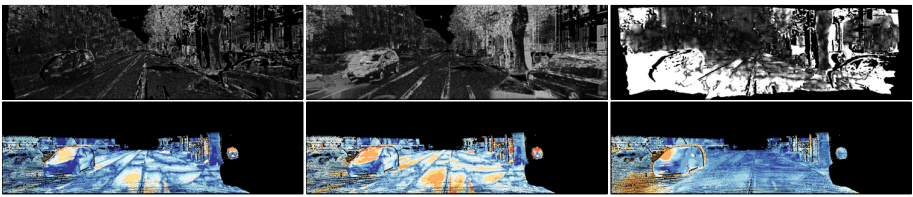}
  \caption{\textbf{Illustration of different consistency terms.} From left to right we show results produced using $\cL_{pd}, \cL_{pf}, \cL_{df}$. The first row presents the pixel-wise loss map of the feedforward prediction, where brighter is larger loss. The second row shows the error map for disparity estimation, when compared with the ground-truth label. Red indicates higher error, whereas blue indicates lower error.}
\label{fig:loss}
\end{figure*}

\subsection{Ablation Studies}
\label{sec:ablate}
We provide further experimental analysis on ablated versions of our model, for further insight into the design choices. To ensure a fair comparison, all models used in this section are trained on FlyingThings3D, and results are reported on KITTI2015 without self-supervised finetuning.

\subsubsection*{Consistency in Refinement Network}

In the refinement network, the consistency loss is used as a guidance to refine the network outputs. The consistency loss mainly consists of three terms: a stereo photometric loss ($\cL_{pd}$), a flow photometric loss ($\cL_{pf}$) and a disparity flow consistency loss ($\cL_{df}$). To understand the role of each loss, we conduct an ablation study where different combinations are used in the refinement network. 

\begin{table}[t]
\begin{center}
\setlength{\tabcolsep}{5.0pt}
\begin{tabular}{c  c  c c  c  c c}
\toprule
$\sfl$ & $\cL$ & $\grad$ &  D1 & D2 & F1 & SF  \\ \midrule
\checkmark &  &  & 
10.40 & 25.15 & 32.05 & 42.43 \\ 
\checkmark & \checkmark &  & 
9.77 & 22.80 & 30.95 & 40.47 \\ 
\checkmark &  & \checkmark & 
7.44 & 20.01 & 28.20 & 36.31 \\ 
\checkmark & \checkmark & \checkmark & 
7.10  & 19.68  & 27.35 &  35.40 \\ \bottomrule
\end{tabular}
\end{center}
\caption{\textbf{Ablation study on the input to the refinement network.} All models are trained on FlyingThings3D, and evaluated on KITTI.}
\label{tab:ablation_input}

\end{table}

We summarize the results in table~\ref{tab:ablation_loss}. When no consistency is used, the refinement network only takes the prediction $\sfl$ as input. This baseline results in a scene flow error of of $42.43$, slightly better than the feedforward result which is $43.01$. This suggests that only $\sfl$ is insufficient in effectively guiding refinement. The error can be reduced by additionally including consistency terms, which supports the efficacy of our guided refinement strategy. The contributions of each loss term to the scene flow error are similar. However, each loss term exhibits different improvements over the three sub-measurements. Notably, $\cL_{pd}$ improves $D1$ measure the most, $\cL_{pf}$ improves $F1$ the most, whereas $\cL_{df}$ reduces the $D2$ error, which is related to both disparity and optical flow. The results are understandable, as each loss intrinsically constraints different outputs. For example, the stereo photometric loss  $\cL_{pd}$ relates stronger to disparity, whereas the flow photometric loss $\cL_{pf}$ better constrains flow.

To further understand the impact of different losses, in fig.~\ref{fig:loss} we visualize refinement results by using only $\cL_{pd}$, $\cL_{pf}$ or $\cL_{df}$. We observe that the photometric loss (the left and middle) is sparse and cannot provide sufficient signal in the textureless regions. As a result, the error in flat regions (such as road) remains large after refinement. On the other hand, the disparity flow consistency term $\cL_{df}$ provides a denser signal, especially in textureless regions, and can reduce errors there. However, it still produces errors on the motion boundaries and in occluded areas, due to the output inconsistency in such regions. 

\subsubsection*{Input to Refinement Network}
In the refinement network, consistency is used to guide the change of scene flow predictions. In doing so, we use the current scene flow prediction $\sfl^t$, the consistency computed on the scene flow $\cL_{cst}(\sfl)$, and its gradient as input $\cL_{cst}(\sfl)$. 

In this study, we study the influence of different inputs on the final performance. For this purpose, we train several versions of the refinement network, with various combinations of input features. As shown in table~\ref{tab:ablation_input},  with only the variable as input, the refinement network achieved $42.43$.  By additionally including the loss and gradient as inputs, a further decrease of $0.96$ and $6.12$ are observed, respectively, which suggests that the gradient is a stronger signal for refinement, compared to the loss. Nevertheless, combining the two results in the best performance in scene flow, of $35.40$ error rate. The results highlight the effectiveness of the gradient feature in the learned refinement network.

\subsubsection*{Online Refinement Strategies}
In this experiment we compare our learned refinement module against alternative design choices. In this work we use a refinement network to predict an update from the loss signal, and to refine the scene flow prediction without updating the model parameters. Alternatively, one can perform online finetuning to update the model parameters, based on the same consistency loss. We refer to this approach as parameter finetuning. Another choice is to change the scene flow prediction directly, without passing through the network, which we refer to as output finetuning~\cite{chen2019self}.

We test the two alternatives with the same consistency loss. The results are summarized in table~\ref{tab:ablation_ft}. As seen from the table, our learned refinement achieves similar performance with parameter finetuning. However, as multiple iterations of forward and backward passes are needed for finetuning network parameters, this approach results in a large run time of $120$ seconds per image. Compared to parameter finetuning, our method is much faster. 

\begin{table}[t]
\begin{center}
\setlength{\tabcolsep}{5.0pt}
\begin{tabular}{c  c  c  c  c c}
\toprule
Refinement Method  &  D1 & D2 & F1 & SF & Time \\ \midrule
Outputs 
& 8.53 & 22.82 & 30.61 & 39.95 & 0.5s \\ 
Parameters 
& 6.88 & 19.47 & 27.69 & 35.49 & 120s \\ 
Learned (ours) 
& 7.10  & 19.68  & 27.35 & 35.40 & 0.8s \\ \bottomrule 
\end{tabular}

\vspace{3mm}
\caption{\textbf{Ablation study of online refinement strategies.} We use different online refinement strategies for scene flow estimation. All models are trained on FlyingThings3D and results are reported on KITTI.}
\label{tab:ablation_ft}
\end{center}
\end{table}

\begin{figure*}[t]
\small
\centering
\input{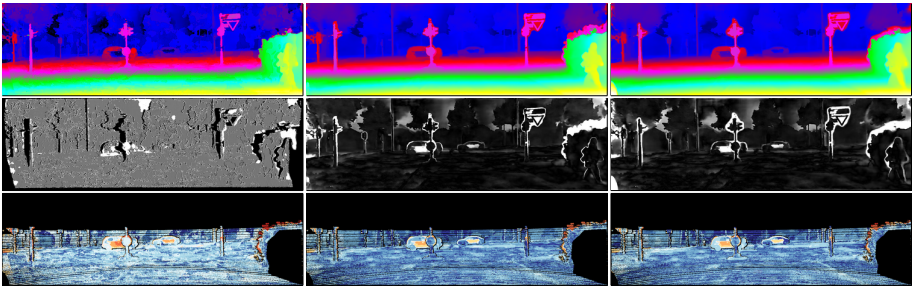}
\caption{\textbf{Qualitative results for different online refinement strategies.} From left to right, we show the results refined with output finetuning, parameter finetuning and our proposed learning-based refinement. The first row illustrates disparity estimation, the second the accumulated updates, where brighter means larger change in disparity. The last row shows the error map evaluated against ground-truth. Red represents higher error, and blue lower error.}
\label{fig:ft}
\end{figure*}

To further understand the difference among the three refinement strategies, we visualize the disparity outputs in fig.~\ref{fig:ft}. We can observe from the figure, that output finetuning is very noisy and sparse -- this is due to properties of each pixel being refined independently. As shown previously, the loss signal is sparse and noisy, and is often invalid in occluded regions, which results in inaccurate updates. In contrast, our learned refinement can take advantage of this sparse signal and of contextual information. As a result, a much smoother refinement--similar to the one produced by parameter finetuning, but \emph{two orders of magnitude faster}--is produced by the proposed learned refinement module.

\subsection{Performance for Unseen Real Visual Data}

\label{sec:zed}
\begin{figure*}[t]
\small
\centering
\input{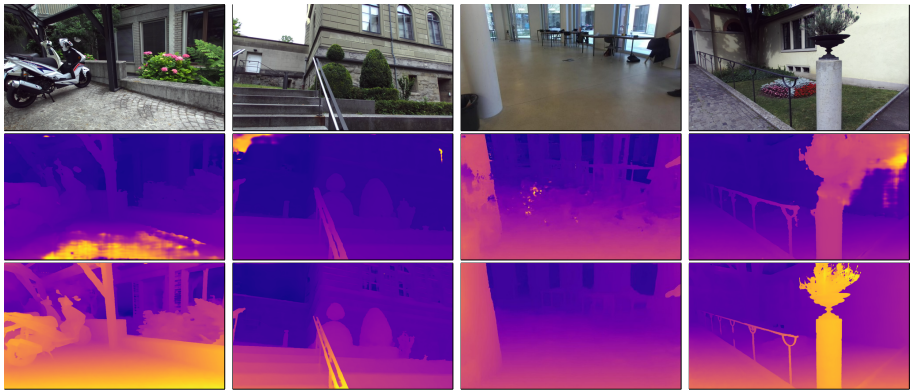}
\caption{\textbf{Disparity estimation results in real scenes different from training.} Top: input left image, Middle: feedforward results, Bottom: results with consistency guided refinement. Note that refinement can significantly improve drastically incorrect predictions.}
\label{fig:zed}
\end{figure*}

To test our method's generalization in other environments, we apply the model to real scenes captured by a ZED stereo camera and compare against results from the feedforward module. As shown in fig.~\ref{fig:zed}, our consistency guided refinement can notably improve the performance of disparity estimation in such cases. This again confirms that scene flow consistency can be a reliable model in overcoming domain shift, and support more robust performance in unseen scenarios. 

\section{Conclusions}
We have presented Consistency Guided Scene Flow (\ours), a framework for the joint estimation of disparity and scene flow from stereo video. 
The consistency loss combines stereo and temporal photo-consistency with a novel geometric model which couples the depth and flow variables. Besides the usual online parameter and output finetuning strategies typical of self-supervised test-time domain adaptation, we introduce new, learned, output finetuning descent models with explicit gradient features, that produce good quality results, on par with parameter finetuning, but two orders of magnitude faster. Extensive experimental validation on benchmarks indicates that \ours{} can reliably predict disparity and scene flow, with good generalization in cross-domain settings.

\clearpage
\bibliographystyle{splncs04}
\bibliography{main_bib}

\begin{thebibliography}{10}
\providecommand{\url}[1]{\texttt{#1}}
\providecommand{\urlprefix}{URL }
\providecommand{\doi}[1]{https://doi.org/#1}

\bibitem{aleotti2020learning}
Aleotti, F., Poggi, M., Tosi, F., Mattoccia, S.: Learning end-to-end scene flow
  by distilling single tasks knowledge. In: AAAI (2020)

\bibitem{andrychowicz2016learning}
Andrychowicz, M., Denil, M., Gomez, S., Hoffman, M.W., Pfau, D., Schaul, T.,
  Shillingford, B., De~Freitas, N.: Learning to learn by gradient descent by
  gradient descent. In: Advances in neural information processing systems. pp.
  3981--3989 (2016)

\bibitem{basha2013multi}
Basha, T., Moses, Y., Kiryati, N.: Multi-view scene flow estimation: A view
  centered variational approach. International journal of computer vision
  \textbf{101}(1),  6--21 (2013)

\bibitem{casser2019depth}
Casser, V., Pirk, S., Mahjourian, R., Angelova, A.: Depth prediction without
  the sensors: Leveraging structure for unsupervised learning from monocular
  videos. In: AAAI (2019)

\bibitem{chang18pyramid}
Chang, J., Chen, Y.: Pyramid stereo matching network. In: CVPR (2018)

\bibitem{chen2019self}
Chen, Y., Schmid, C., Sminchisescu, C.: Self-supervised learning with geometric
  constraints in monocular video: Connecting flow, depth, and camera. In: ICCV
  (2019)

\bibitem{clark2018ls}
Clark, R., Bloesch, M., Czarnowski, J., Leutenegger, S., Davison, A.J.: Ls-net:
  Learning to solve nonlinear least squares for monocular stereo. arXiv
  preprint arXiv:1809.02966  (2018)

\bibitem{dosovitskiy2015flownet}
Dosovitskiy, A., Fischer, P., Ilg, E., Hausser, P., Hazirbas, C., Golkov, V.,
  Van Der~Smagt, P., Cremers, D., Brox, T.: {FlowNet}: Learning optical flow
  with convolutional networks. In: ICCV (2015)

\bibitem{godard2017unsupervised}
Godard, C., Mac~Aodha, O., Brostow, G.J.: Unsupervised monocular depth
  estimation with left-right consistency. In: CVPR (2017)

\bibitem{godard2019digging}
Godard, C., Mac~Aodha, O., Firman, M., Brostow, G.J.: Digging into
  self-supervised monocular depth estimation. In: ICCV (2019)

\bibitem{guo2018learning}
Guo, X., Li, H., Yi, S., Ren, J., Wang, X.: Learning monocular depth by
  distilling cross-domain stereo networks. In: ECCV (2018)

\bibitem{huguet2007variational}
Huguet, F., Devernay, F.: A variational method for scene flow estimation from
  stereo sequences. In: ICCV (2007)

\bibitem{ilg2017flownet}
Ilg, E., Mayer, N., Saikia, T., Keuper, M., Dosovitskiy, A., Brox, T.:
  {FlowNet} 2.0: Evolution of optical flow estimation with deep networks. In:
  CVPR (2017)

\bibitem{Ilg2018occlusions}
Ilg, E., Saikia, T., Keuper, M., Brox, T.: Occlusions, motion and depth
  boundaries with a generic network for disparity, optical flow or scene flow
  estimation. In: ECCV (2018)

\bibitem{jiang2019sense}
Jiang, H., Sun, D., Jampani, V., Lv, Z., Learned-Miller, E., Kautz, J.: Sense:
  A shared encoder network for scene-flow estimation. In: ICCV (2019)

\bibitem{Kendall_2017_ICCV}
Kendall, A., Martirosyan, H., Dasgupta, S., Henry, P., Kennedy, R., Bachrach,
  A., Bry, A.: End-to-end learning of geometry and context for deep stereo
  regression. In: ICCV (2017)

\bibitem{kuznietsov2017semi}
Kuznietsov, Y., Stuckler, J., Leibe, B.: Semi-supervised deep learning for
  monocular depth map prediction. In: CVPR (2017)

\bibitem{lai2019bridging}
Lai, H.Y., Tsai, Y.H., Chiu, W.C.: Bridging stereo matching and optical flow
  via spatiotemporal correspondence. In: CVPR (2019)

\bibitem{li2016learning}
Li, K., Malik, J.: Learning to optimize. arXiv preprint arXiv:1606.01885
  (2016)

\bibitem{liu2019selflow}
Liu, P., Lyu, M., King, I., Xu, J.: Selflow: Self-supervised learning of
  optical flow. In: CVPR (2019)

\bibitem{lv2019taking}
Lv, Z., Dellaert, F., Rehg, J.M., Geiger, A.: Taking a deeper look at the
  inverse compositional algorithm. In: CVPR (2019)

\bibitem{Ma2019DRISF}
Ma, W.C., Wang, S., Hu, R., Xiong, Y., Urtasun, R.: Deep rigid instance scene
  flow. In: CVPR (2019)

\bibitem{mayer16a}
Mayer, N., Ilg, E., H{\"{a}}usser, P., Fischer, P., Cremers, D., Dosovitskiy,
  A., Brox, T.: A large dataset to train convolutional networks for disparity,
  optical flow, and scene flow estimation. In: CVPR (2016)

\bibitem{meister2018unflow}
Meister, S., Hur, J., Roth, S.: {UnFlow}: Unsupervised learning of optical flow
  with a bidirectional census loss. In: AAAI (2018)

\bibitem{Menze2015ISA}
Menze, M., Heipke, C., Geiger, A.: Joint 3d estimation of vehicles and scene
  flow. In: ISPRS Workshop on Image Sequence Analysis (ISA) (2015)

\bibitem{Menze2018JPRS}
Menze, M., Heipke, C., Geiger, A.: Object scene flow. ISPRS Journal of
  Photogrammetry and Remote Sensing (JPRS)  (2018)

\bibitem{pang2018zoom}
Pang, J., Sun, W., Yang, C., Ren, J., Xiao, R., Zeng, J., Lin, L.: Zoom and
  learn: Generalizing deep stereo matching to novel domains. In: CVPR (2018)

\bibitem{poggi2019guided}
Poggi, M., Pallotti, D., Tosi, F., Mattoccia, S.: Guided stereo matching. In:
  CVPR (2019)

\bibitem{sun2018pwc}
Sun, D., Yang, X., Liu, M.Y., Kautz, J.: {PWC-net}: {CNN}s for optical flow
  using pyramid, warping, and cost volume. In: CVPR (2018)

\bibitem{tang2018ba}
Tang, C., Tan, P.: Ba-net: Dense bundle adjustment network. arXiv preprint
  arXiv:1806.04807  (2018)

\bibitem{tonioni2017unsupervised}
Tonioni, A., Poggi, M., Mattoccia, S., Di~Stefano, L.: Unsupervised adaptation
  for deep stereo. In: ICCV (2017)

\bibitem{tonioni2019learning}
Tonioni, A., Rahnama, O., Joy, T., Stefano, L.D., Ajanthan, T., Torr, P.H.:
  Learning to adapt for stereo. In: CVPR (2019)

\bibitem{tonioni2019real}
Tonioni, A., Tosi, F., Poggi, M., Mattoccia, S., Stefano, L.D.: Real-time
  self-adaptive deep stereo. In: CVPR (2019)

\bibitem{vedula1999three}
Vedula, S., Baker, S., Rander, P., Collins, R., Kanade, T.: Three-dimensional
  scene flow. In: ICCV (1999)

\bibitem{vogel2013piecewise}
Vogel, C., Schindler, K., Roth, S.: Piecewise rigid scene flow. In: ICCV (2013)

\bibitem{wang2019unos}
Wang, Y., Wang, P., Yang, Z., Luo, C., Yang, Y., Xu, W.: Unos: Unified
  unsupervised optical-flow and stereo-depth estimation by watching videos. In:
  CVPR (2019)

\bibitem{wang2004image}
Wang, Z., Bovik, A.C., Sheikh, H.R., Simoncelli, E.P., et~al.: Image quality
  assessment: from error visibility to structural similarity. IEEE transactions
  on image processing  \textbf{13}(4),  600--612 (2004)

\bibitem{wedel2008efficient}
Wedel, A., Rabe, C., Vaudrey, T., Brox, T., Franke, U., Cremers, D.: Efficient
  dense scene flow from sparse or dense stereo data. In: ECCV (2008)

\bibitem{yin2018geonet}
Yin, Z., Shi, J.: {GeoNet}: Unsupervised learning of dense depth, optical flow
  and camera pose. In: CVPR (2018)

\bibitem{zhang2019ga}
Zhang, F., Prisacariu, V., Yang, R., Torr, P.H.: Ga-net: Guided aggregation net
  for end-to-end stereo matching. In: CVPR (2019)

\bibitem{zhong2018open}
Zhong, Y., Li, H., Dai, Y.: Open-world stereo video matching with deep rnn. In:
  ECCV (2018)

\bibitem{zhou2017unsupervised}
Zhou, T., Brown, M., Snavely, N., Lowe, D.G.: Unsupervised learning of depth
  and ego-motion from video. In: CVPR (2017)

\end{thebibliography}

\clearpage

\section*{Appendix}
In the appendix, we provide an analysis of the iterative refinement module. 

The consistency-guided refinement module is used to iteratively refine the scene flow estimation. To gain further understanding of the refinement process, we provide an analysis on the intermediate results, \ie, scene flow estimation after each refinement step.

\begin{figure*}[t]
\small
\centering
\includegraphics[trim=80 180 80 150, clip, width=0.8\linewidth]{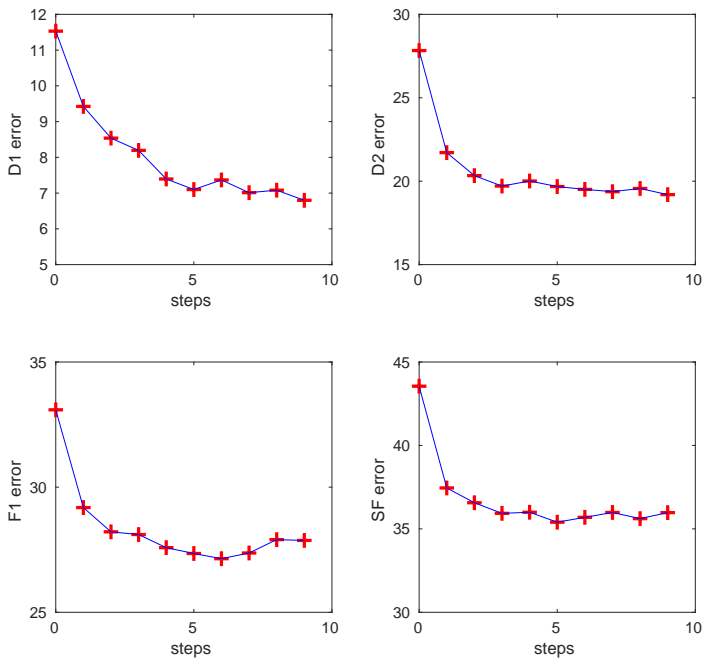}
\caption{\textbf{The effect of refinement steps on scene flow performance.} We plot the error using D1, D2, F1 and SF measure in the above four figures, respectively. x-axis is the number of refinement steps, with $0$ being the feedforward results(without the refinement module). y-axis is the percentage of outliers.  A pixel is considered correct if the prediction end-point error is smaller than $3px$ or $5\%$. For SF, a correct pixel needs to be correct for D1, D2 and F1. The model is trained on FlyingThings3D, and test on KITTI scene flow dataset.}
\label{fig:sf_error_iter}
\end{figure*}

Following the main paper, we use the same experimental setting in this study. In more details, the model is trained on the FlyingThings 3D dataset using ground-truth labels, and evaluated on the KITTI scene flow dataset. We evaluate the scene flow estimation results after each refinement step, using the four evaluation metrics: D1, D2, F1, SF, which represent the percentage of outliers. In the main paper, a total of 5 refinement steps are performed. Here we also report the performance of applying further refinement, up to $10$ steps. We plot the results at different steps in fig~\ref{fig:sf_error_iter}. We observe that the error decreases notably in all four measurements, which shows that the refinement module is useful for improving scene flow accuracy. The improvement is more significant in the first $5$ steps, then the performance saturates with additional steps. Further refinement leads to an increase in the overall running time, as per refinement step requires approximately 0.1 second. 

\begin{figure*}[t]
\small
\centering
\input{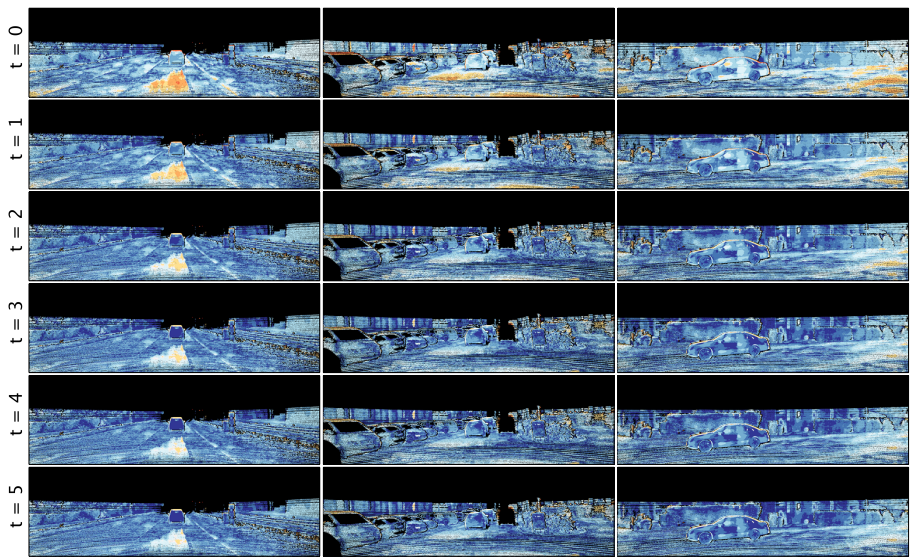}
\caption{\textbf{Error map in different refinement step.} We show the error map of disparity estimation. Blue represents lower error, while red represents higher error. Note that the refinement module produces in a considerable decrease in the error.}
\label{fig:iters}
\end{figure*}

We present some pixel-wise error visualizations in fig~\ref{fig:iters}, where we show the error maps at different refinement steps. We can observe that the proposed refinement module also produces a notable qualitative improvement in every iteration, this further verifies the effectiveness of the consistency-guided refinement. 

\end{document}